\documentclass[nohyperref]{article}

\usepackage{microtype}
\usepackage{graphicx}

\usepackage{amsmath,amsfonts}
\usepackage{array}
\usepackage{textcomp}
\usepackage{stfloats}
\usepackage{url}
\usepackage{verbatim}
\usepackage{vcell}
\usepackage{booktabs}
\usepackage{multirow}
\usepackage{hhline}
\usepackage{caption}
\usepackage{subfigure}
\usepackage{enumitem}

\newcommand*{\oursfm}{FollowMe-STGCNN }

\usepackage{hyperref}

\usepackage[accepted]{icml2022}

\usepackage{amssymb}
\usepackage{mathtools}
\usepackage{amsthm}

\usepackage[capitalize,noabbrev]{cleveref}

\theoremstyle{plain}

\theoremstyle{definition}

\theoremstyle{remark}

\usepackage[textsize=tiny]{todonotes}

\icmltitlerunning{FollowMe: Vehicle Behaviour Prediction in Autonomous Vehicle Settings}
\begin{document}

\twocolumn[
\icmltitle{FollowMe: Vehicle Behaviour Prediction in Autonomous Vehicle Settings}



\icmlsetsymbol{equal}{*}

\begin{icmlauthorlist}
\icmlauthor{Abduallah Mohamed}{equal,yyy,ttt}
\icmlauthor{Jundi Liu}{equal,www}
\icmlauthor{Linda Ng Boyle}{www}
\icmlauthor{Christian Claudel}{ttt}
\end{icmlauthorlist}

\icmlaffiliation{yyy}{Meta Reality Labs, work was done before joining Meta Reality Labs.}
\icmlaffiliation{www}{University of Washington}
\icmlaffiliation{ttt}{The University of Texas at Austin}

\icmlcorrespondingauthor{Abduallah Mohamed}{abduallahmohamed.com}


\vskip 0.3in
]



\printAffiliationsAndNotice{\icmlEqualContribution} 

\begin{abstract}
An ego vehicle following a virtual lead vehicle planned route is an essential component when autonomous and non-autonomous vehicles interact. Yet, there is a question about the driver's ability to follow the planned lead vehicle route. Thus, predicting the trajectory of the ego vehicle route given a lead vehicle route is of interest. We introduce a new dataset, the FollowMe dataset, which offers a motion and behavior prediction problem by answering the latter question of the driver's ability to follow a lead vehicle. We also introduce a deep spatio-temporal graph model FollowMe-STGCNN as a baseline for the dataset. In our experiments and analysis, we show the design benefits of FollowMe-STGCNN in capturing the interactions that lie within the dataset. We contrast the performance of FollowMe-STGCNN with prior motion prediction models showing the need to have a different design mechanism to address the lead vehicle following settings.
\end{abstract}

\section{Introduction}

The coming era of interaction between autonomous and non-autonomous vehicles requires equipping non-autonomous vehicles with guidance capabilities to interact with the autonomous ones. The autonomous vehicles can communicate with each other and create a cooperative space-time planned routes to cross intersections and roads without stopping, minimizing travel times. The existence of non-autonomous agents can impair such an advantage due to the lack of coordination with the autonomous entities in the routes planning. One solution for this problem is to equip the non-autonomous vehicles with a virtual display that shows a virtual lead vehicle to follow. The driver will follow the lead vehicle planned route which is coordinated with the autonomous vehicles routes. The lead vehicle displays the route dynamically by respecting the space-time aspects where it speeds up or slows down to have the driver accelerate or decelerate based on the planned route. One major issue is how to plan such routes that are natural and realistic for the drivers to follow. For example, some drivers when they turn right they might create a virtual lane that is between the first and second lane because it is easier for them to do so.
\begin{figure}[t]
    \centering
    \includegraphics[width=\columnwidth]{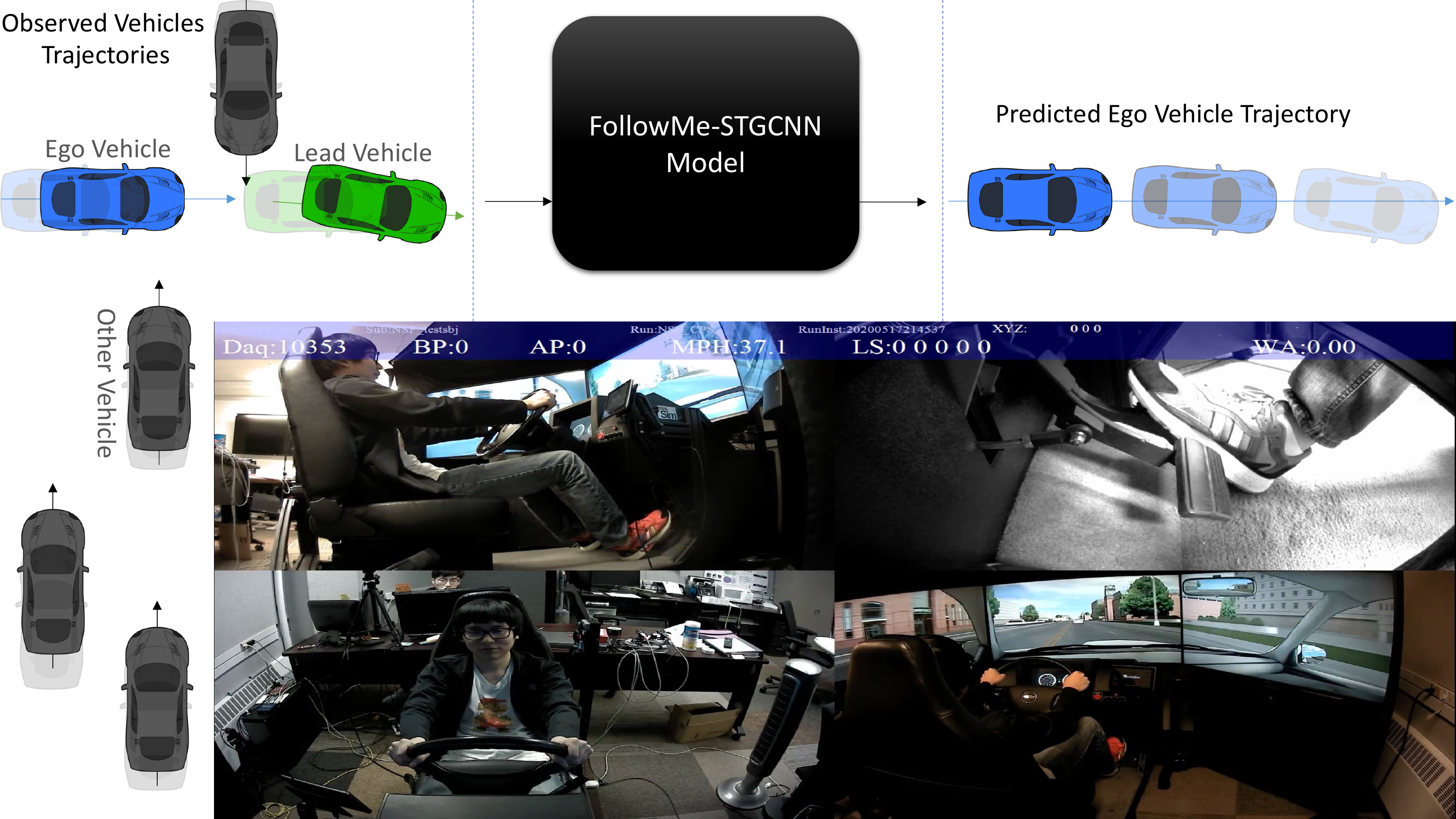}
    \caption{We introduce a new dataset for the lead vehicle following challenge. Our introduced baseline model \oursfm models the observed interactions between the ego, lead and other vehicles to predict the future trajectory of the ego vehicle. We attempt to answer the question of how well the ego vehicle will follow the lead vehicle.}
    \label{fig:DatasetTeaser}
\end{figure}
Some drivers like to be close to the center of the lane while others might prefer the edge of the lane. Other cases when a truck is coming in the other direction, some defensive drivers might slightly tilt away from it for safety reasons. Thus, planning a natural and realistic route is quite challenging. Thus having the ability to predict the ego vehicle trajectory when following a lead vehicle is vital for this motion planning problem. If one is able to predict what will the driver do given a lead vehicle route he can adjust the lead vehicle route to have a better following behavior. In this work we introduce a new dataset, FollowMe dataset. This dataset is the result of a study conducted to assess the behaviors of drivers following a lead vehicle. The study was conducted over 32 drivers in different driving scenarios with a hardware emulation that gives a realistic experience. The ego vehicle’s route driven by the drivers, the lead vehicle route and other vehicles routes in the environment were recorded. This dataset forms a new motion prediction problem. The problem statement is as follows, given an observation of ego vehicle route, lead vehicle route and other vehicles routes in the environment what will be the future trajectory of the ego vehicle. Answering this question will allow the lead vehicle route designers to create better routes that will maximize the drivers ability to follow the lead vehicle.

Beside the study and the generated dataset we introduce a new deep spatio-temporal graph model \oursfm that acts as a baseline for this dataset for future research directions. \oursfm model is tailored to the target of this problem which is predicting the future trajectory of the ego vehicle when it follows a lead vehicle. We can consider the existence of the lead vehicle in the motion prediction problem as a systematic bias in the dataset that needs to be addressed to predict the ego vehicle trajectory. Previous motion prediction models were tailored for interaction between agents without the existence of such a bias. Beside this bias, these motion forecasting deep models tend to predict the whole scene, something that leads to inaccurate predictions when used in our settings because the prediction target would be overloaded with predicting multiple targets. Thus, we show that some design aspects in the deep model we propose, such as the usage of attention can reflect this bias.  In our case, \oursfm is designed in a way that focuses only on the ego vehicle prediction. We analyze our model quantitatively and qualitatively and show that our introduced baseline model indeed captures the important aspects and behaviors of the data unlike prior works. The contribution of this paper as follows:
\setlist{nosep,after=\vspace{\baselineskip}}
\begin{itemize}[noitemsep]
    \item Introduction of a new dataset for lead vehicle following problem.
    \item Introduction of a new motion prediction problem that is biased by the lead vehicle.
    \item Introduction of a new baseline deep model, \oursfm that is tailored for this problem.
\end{itemize}
In the next sections we will review the prior works related to our work, introduce the study and the dataset, describe our model and the training mechanism. Then we finish with extensive qualitative and quantitative experiments that analyze and benchmark our model in comparison with multiple baselines with inner insights of our model's behavior.
\section{Literature Review}
\subsection{Guidance Visualization and Related Datasets}
To better assist the non-autonomous vehicles to cooperate with the autonomous vehicles in the near future, researchers have proposed multiple guidance presentation methods in the driver-vehicle interface. Wang \textit{el al.}~\cite{wang2020augmented} adopted an augmented reality (AR)-based slot reservation system to help the drivers visualize the guidance information while driving through an unsignalized intersection for connected vehicles. The guidance was shown as colored lanes on the windshield. Other studies also investigated the AR-based information presentation including navigation arrows, bounding boxes for the scene objects, etc~\cite{liu2021clustering, akash2020toward}. The datasets used in these studies are mainly non-open source. However, we proposed to present the guidance information by introducing a virtual lead vehicle which can better demonstrate the desired speed and trajectory for drivers to follow. Moreover, to better assist researchers investigate on this topic, our dataset is fully open-sourced.

\subsubsection{Vehicle Following Datasets}
Systematically analyzing the vehicle following problem is crucial to design such lead vehicles. Recent advancement on vehicle following models are mainly focusing on deep learning models \cite{zhang2019simultaneous, zhu2018human, tang2020car}.

Table~\ref{tab:car_following_datasets} shows the comparison of driving datasets that have been used for training and testing the vehicle following models. Next Generation SIMulation (NGSIM)~\cite{NGSIM} program launched by U.S. Department of Transportation Federal Highway Administration provided two highway driving datasets (US 101 and I-80) that contains vehicle following data.
The NGSIM datasets are naturalistic driving datasets with limited study areas and short duration of vehicle following information for a specific vehicle. Moreover, the driving operations are limited to follow a lead vehicle to drive straightly with less sample frequency. As for Shanghai Naturalistic Driving Study (SH-NDS)~\cite{zhu2018human}, it has better variety of driving operations in an urban environment and higher sample frequency. However, the vehicle following duration for a specific driver is still shorter than our proposed FollowMe dataset. Our dataset has a highest sampling rate, longest segment, and relatively large number of scenes and total length of data collection.

\begin{table*}[t]
\tiny
\caption{Comparison of datasets used for vehicle following problem.}
\label{tab:car_following_datasets}
\begin{tabular}{lllll}
\toprule
 &\textbf{FollowMe(ours)}& US 101(NGSIM) & I-80(NGSIM) & SH-NDS  \\ \hline
Traffic type & Autonomous and non-autonomous& Non-autonomous & Non-autonomous & Non-autonomous  \\
Other traffic direction &Multiple& Single & Single & Multiple   \\
Lead vehicle type & Controlled autonomous & Uncontrolled non-autonomous & Uncontrolled non-autonomous & Uncontrolled non-autonomous \\
Study area &$\sim$5000 meters (16,400 feet)& $\sim$640 meters (2,100 feet) & $\sim$500  meters  (1,640  feet) & N/A  \\
Vehicle following duration of a single car &$\sim$50 mins& $\sim$30 secs & $\sim$20 secs & \textgreater 15secs   \\
Driving environment &Suburban& Highway & Highway & Urban  \\
Driving operations &Straight, left and right turns& Straight & Straight & Straight, left and right turns  \\
Frequency &60Hz& 10 Hz & 10 Hz & 10-50 Hz  \\
Data sensory range & All in environment& 100m & 100m & 120m \\ \bottomrule
\end{tabular}
\end{table*}

\subsubsection{Motion Prediction Datasets}

Motion prediction methodologies can be applied to multiple use cases. For example, predicting pedestrians' trajectories use real-world human trajectories with rich human-human interaction scenarios. Such datasets include ETH \cite{eth_biwi_00534} and the more recent ActEV/VIRAT \cite{awad2018trecvid}. Interactive motion prediction for autonomous driving using Waymo Open Motition dataset \cite{ettinger2021large}. A comparison of related datasets are shown in Table \ref{tab:datasets_motion_pred}. Our dataset has mixed autonomous and non-autonomous traffic, higher segment length for a single driver, and a higher sample rate. The high sample rate can help us capture the more detailed temporal-spacial changes of the trajectories given the high speed of the vehicles.

\begin{table*}[t]
\scriptsize
\centering
\caption{Comparison of datasets used for motion prediction.}
\label{tab:datasets_motion_pred}
\begin{tabular}{lllll}
\toprule
 & \textbf{FollowMe(ours)}& ETH & ActEV/VIRAT & Waymo Open Motion  \\ \hline
Traffic Type & Autonomous and Non-autonomous & Non-autonomous & Non-autonomous & Non-autonomous  \\ 
Scene Objects &Vehicles& Pedestrians and vehicles & Pedestrians and vehicles & Pedestrians, cyclists and vehicles   \\
Segment length &50 mins& 60 secs & N/A & 20 secs   \\
Total lengh & 30 hrs& 5 mins & 12 hrs & 574 hrs  \\
\# of scenes & 12& 3 & 12 & 6  \\
Sampling rates &60 Hz& 13-14 Hz & 30 Hz & 10 Hz \\ \bottomrule
\end{tabular}
\end{table*}

\subsection{Motion Prediction Approaches}
Data driven motion prediction proved to be better in performance when compared to traditional methods~\cite{gupta2018social}. The earlier works of~\cite{gupta2018social,alahi2016social} proposed an early version deep models for motion predictions with an emphasis on the social interaction between agents. They highlighted social interactions as a core component to enhance motion prediction models. Yet, the method in modeling the interactions was based on a pooling mechanism and recurrent deep models which resulted in information loss. Later one, the works of~\cite{mohamed2020social,huang2019stgat,li2020evolvegraph} showed that modeling the motion prediction problem as a spatio-temporal graph by using graph CNNs~\cite{kipf2016semi} lead to better performance. They offered to model the interactions using the graph edges directly~\cite{mohamed2020social,li2020evolvegraph} or by using an internal attention mechanism~\cite{huang2019stgat,haddad2021self}. Other methods started using the graph based approach with an egocentric view of each agent~\cite{mangalam2020not,salzmann2020trajectron++} leading to a better performance when compared with prior approaches. Later on, recent deep models treated the problem by adding an extra component that predicts the final goal~\cite{mangalam2020not,zhao2021you}. Yet, none of these prior works had the current settings in which a vehicle is following a lead one. This lead vehicle following settings can be understood as a systematic bias in the dataset in which a proper modeling is needed. This is where our proposed base model \oursfm fill this gap and opens future research directions in modeling such a problem.

\section{Dataset Description}
\begin{figure*}
    \centering
    \includegraphics[width=\linewidth]{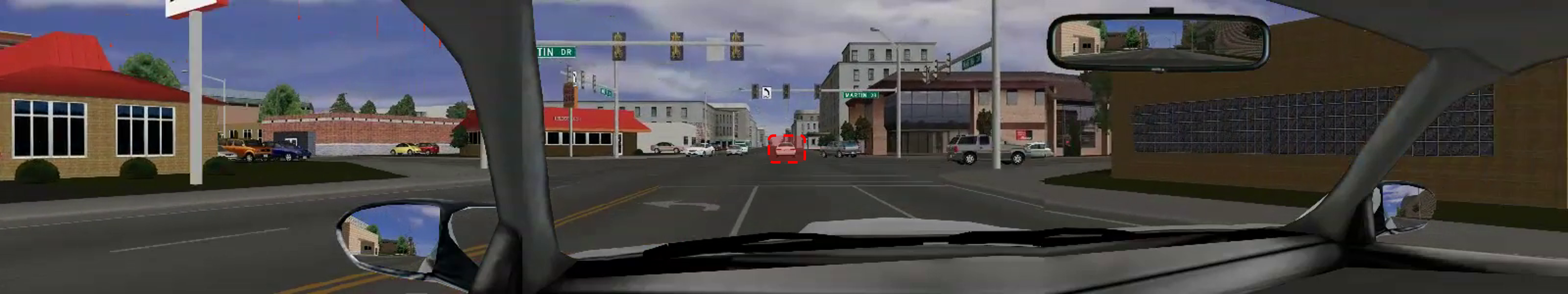}
    \caption{Sample from the simulator. The driver is driving through a busy intersection following the lead vehicle marked with the red box.}
    \label{fig:3DVisDataset}
\end{figure*}

We introduce the FollowMe dataset. This dataset is a result of a study that was created to assess drivers behaviors when following a lead vehicle. 
The driving simulator study used a simulated suburban environment for collecting driver behavior data while following a lead vehicle driving through various unsignalized intersections. The driving environment had two lanes in each direction and a dedicated left-turn lane for the intersection. The experiment was deployed on a miniSim Driving Simulator.
Four video recording units capturing the driver's behavior simultaneously were shown in Figure~\ref{fig:DatasetTeaser}. During the study, the participants were instructed to follow a lead vehicle driving through five intersections (three driving straight intersections, one left turn intersection and one right turn intersection) to maintain a safe distance between 50 feet to 200 feet. An example figure of driver's view was shown in Figure~\ref{fig:3DVisDataset}.
Three within-subject factors were varied for the five intersections: traffic density, speed of the lead vehicle, and operation types. The traffic density had two levels: with or without competing traffic. This promoted for understanding the effect of other agents on the driver's behavior. The speed of the lead vehicle had two levels: 30 and 40 mph. The two levels were chosen based on the common speed limit for a two-lane suburban environment. The speed of the lead vehicle helped in understanding the effect of the drivers following speed. Lastly, the operations in the intersections have three types: going straight, turning right, and turning left. We programmed the lead vehicle randomly turn left, right or go straight through the intersections. In total we had $2 \text{(traffic density)} \times 2 \text{(speed level)} \times 3 = 12$ scenarios for each participant to be tested on.

\emph{Participants: }Thirty two participants (21 males and 11 females) with ages between 19 and 69 years (mean: 32 years old) participated in and completed the study. The participants were recruited and screened using an online questionnaire. The participants were required to have a valid driver's license for more than two years, drive at least once per week, and have not participated in any driving simulator study in the past six months. 

The compensation was \$40 for their participation, and each participant provided their written consent. The Institutional Review Board at the University 
approved the study. Each participant completed all aforementioned 12 scenarios. Before the participants began the main study, they were given brief instructions about the study, and they completed a tutorial consisting of three intersections that helped familiarize them with the operations of the driving simulator. The entire study took approximately one hour to complete.

To prepare the data for the motion prediction problem, we split the data into train, validation and test sets. The setup for the task is to observe 1 second of trajectories and predict the next 3,5 and 8 seconds. This setting is the same as Waymo~\cite{sun2020scalability} dataset used for vehicle motion prediction.

\section{FollowMe-STGCNN Model}
\textbf{Problem Formulation:} We are given an observation of ego, lead and other Vehicles positions $p= (x,y), p \in P$ over an observation time $T_o$ where the subscript $o$ denotes observation. The class of vehicles $c \in C$ is ego, lead or other. The number of observed vehicles is $N$. Thus, we observe an $N$ vehicle with $[P,C]$ position and class attributes over time span $T_o$. The goal is to predict the future trajectory of the ego vehicle over a time span $T_p$, where $p$ subscript denotes prediction. We form a spatio-temporal graph for each observed agent$\mathcal{V}_n$, where the set of all observations is $tr_{o}$ with the dimensions of $[P,C]\times T_o \times N$ as an input to our model. Each node $n \in N$ represents an observed agent $\mathcal{V}_n$ with dimensions of $[P,C]\times T_o$. The target of our model is $\mathcal{V}_{\text{ego}}$ with dimensions of $P \times T_p$. Our model comprises three components. The per node spatial-temporal processor, the spatial-class embedding, and the fusion block. These components are discussed in the coming subsections.

\begin{figure*}[ht]
    \centering
    \includegraphics[width=0.85\linewidth, keepaspectratio]{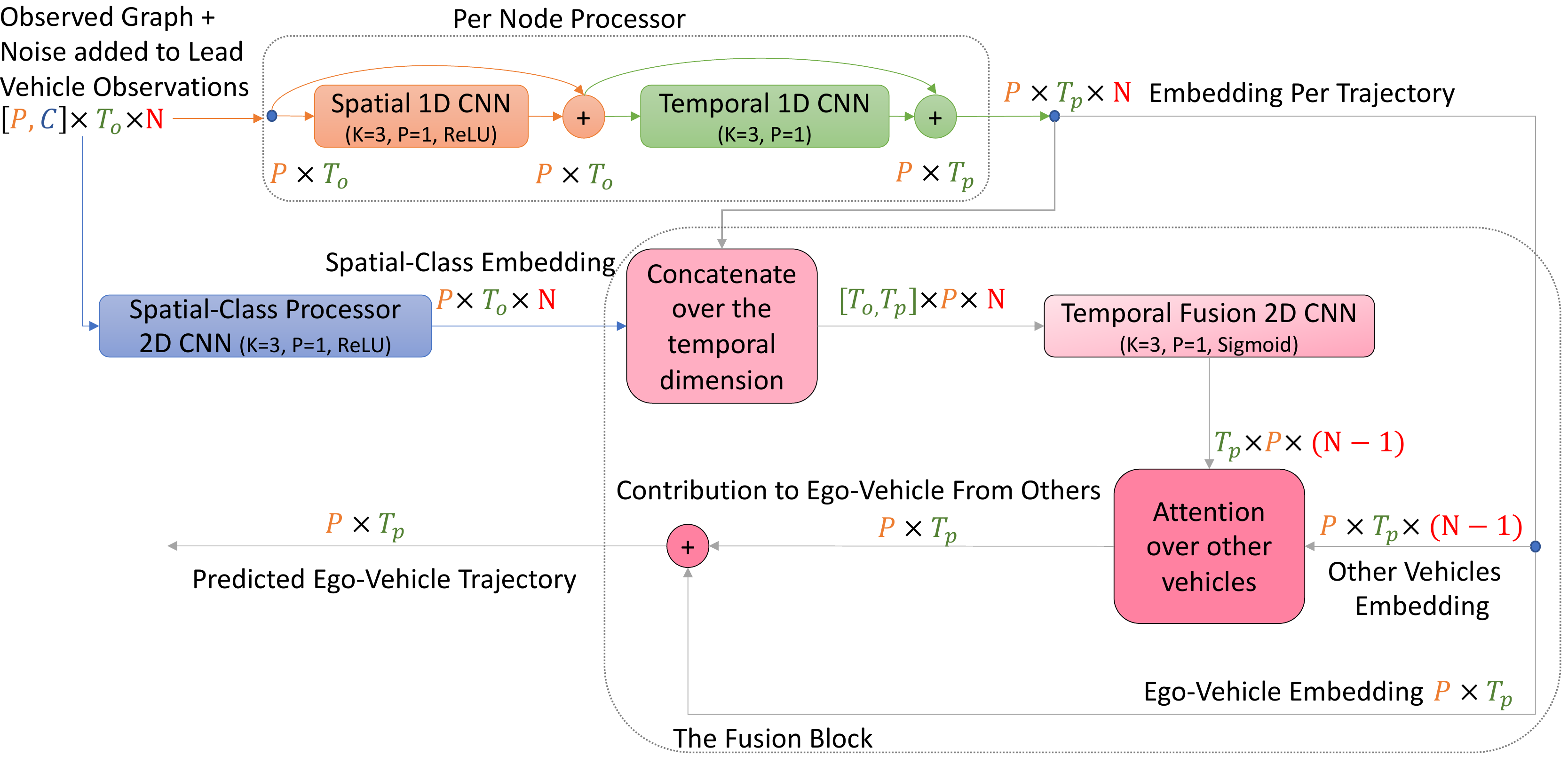}
    \caption[\oursfm model description]{\oursfm model description. Mainly, the per trajectory processor creates an embedding for each observed trajectory. A spatial-class CNN creates a global embedding of the trajectories and their classes that are fused with single trajectories embedding. This fused representation is used as an attention mechanism to weight the contribution of the lead and other vehicles to the embedding of the ego vehicle predicted trajectory. In blocks, K is the kernel size and P is the padding size.}
    \label{fig:followmestgcnnmodel}
\end{figure*}

\textbf{The per node processor:} This part treats each graph node, which is an observed trajectory separately. Each graph node contains spatial and temporal information. We first process the spatial information using a CNN that is illustrated in Figure~\ref{fig:followmestgcnnmodel} treating the spatial features as a features channel. Then we use a temporal CNN that treats the time dimensions as a feature channel. The output of this step is an embedding $\text{Emb}_{n}$ that represents each observed agent trajectory separately with dimensions of $P\times T_p \times N$.
\begin{equation}
    \text{Emb}_{n} = \text{PerNodeProcesser}(\mathcal{V}_n)
\end{equation}
\textbf{The spatial-class processor:} This model treats all the graph nodes. It attempts to extract a global useful feature that utilises both observed trajectories and their observed class. This allows the model to produce an embedding of spatio-temporal data while considering the class of the vehicle. The output of this layer is an embedding $\text{Emb}_{\text{SpatialClass}}$ with dimensions of $P\times T_o \times N$.
\begin{equation}
    \text{Emb}_{\text{SpatialClass}} = \text{SpatialClassCNN}(tr_{o})
\end{equation}
\textbf{The fusion block:} This component defines the contribution of all vehicles to the lead vehicle. As we noticed, our input $tr_{o}$ comes out without a pre-defined connection between the graph's nodes. Instead, in the fusion block, we discover this relationship via the fusion block. The first part concatenates both the spatial-class embedding $P\times T_o \times N$ and the per node embedding  $P\times T_p \times N$ into a single representation $\text{Emb}_{\text{Cat}} $ with dimensions of $[T_o, T_p] \times P \times  N$. We chose to concatenate over the temporal dimension, not to distort the spatial embedding from the per node processor.
\begin{equation}
    \text{Emb}_{\text{Cat}} = \text{Concatenate}(\text{Emb}_{\text{SpatialClass}},\text{Emb}_{N} )
\end{equation}
Where $\text{Emb}_{N}$ is the concatenation of all $\text{Emb}_{n}$ embeddings.
This single representation is now passed through a fusion CNN that produces a weighting tensor between 0 and 1. We chose it be in this range of 0-1 to have a range of representation reflecting no contribution to a range of full contribution. Thus, the Sigmoid activation function was used in this context. Our goal is to figure the relationship between the ego-vehicle and others, the dimension of the fusion CNN output $\mathcal{F}_\text{FusionWeighting}$ is of $P \times T_p \times (N-1) \times 1$. We notice that the $N$ dimensions are less by one, because we want to use the lead and other vehicle features to enhance the ego vehicle trajectory prediction without repeating the information from the ego vehicle embedding. This fusion output determines the weight of contribution from lead and other vehicles to the future trajectory of the ego-vehicle. We can think of it in terms of query and key that is used in regular attention mechanism, where the query is the $\text{Emb}_{\text{SpatialClass}}$ and the key is the $\text{Emb}_{N}$.
\begin{equation}
    \mathcal{F}_\text{FusionWeighting} = \text{Sigmoid}(\text{CNN}(\text{Emb}_{\text{Cat}}))
\label{eq:fusion_weight}
\end{equation}

We now use an attention mechanism to determine this contribution from lead and other vehicles into the future ego vehicle trajectory. We perform the attention using the $\mathcal{F}_\text{FusionWeighting}$ and the single embedding of lead and other vehicles $\text{Emb}_{N-1}$ without the ego vehicle embedding. This attention $\text{ATTN}$ is in dimensions of $P\times T_p$. In other terms, we attempt to weight each lead and other vehicles embedding $\text{Emb}_{N-1}$. In the experiments section, we analyze the behavior of the fusion block regarding the lead and other vehicles.
\begin{equation}
    \text{ATTN}=\text{MatMul}(\mathcal{F}_\text{FusionWeighting}, \text{Emb}_{N-1})
\label{eq:attn_output}
\end{equation}
The output of the attention carries the contribution from other vehicles to the ego-vehicle. For the final output, we add the $\text{ATTN}$ to the embedding of the ego vehicle $\text{Emb}_{\text{ego}}$ resulting in the predicted trajectory with dimensions of $P\times T_p$.
\begin{equation}
    \text{Ego vehicle predicted trajectory}=\text{ATTN} + \text{Emb}_{\text{ego}}
\label{eq:final_output}
\end{equation}

\section{Training Mechanism and Loss Function}
Our model is a generative model that generates a distribution of the predicted ego vehicle trajectory as trajectory samples. There are plenty of methods to train generative models starting from Maximum Likelihood Estimation (MLE), Variational AutoEncoders (VAEs)~\cite{kingma2013auto} and Generative Adversarial Networks (GANs)~\cite{goodfellow2014generative}. MLE requires a lot of samples to converge. VAEs and GANs require special design of the network. The latter two methods, besides being difficult to stabilize, are prone to mode dropping~\cite{li2018implicit}. A straightforward generative training mechanism called Implicit Maximum Likelihood Estimation (IMLE) was introduced by~\cite{li2018implicit} to overcome this issue. The goal of this method is to generate samples that are close to the ground truth, mimicking the drivers' behaviors by forcing each generated sample to be close to a ground truth trajectory. IMLE works by injecting a noise into the model input, sampling multiple samples and selecting the one closest to ground truth and back propagating from there using the loss function. IMLE training is stable because the aim only focuses on the generated samples, unlike GANs, which require a generator-discriminator objective and VAEs which optimizes Evidence Lower Bound (ELBO). For more details about IMLE, the original paper by~\cite{li2018implicit} contains more elaboration. In our settings, we modify the IMLE in which we condition the noise input to the model with the lead vehicle's observed trajectory. The training mechanism of lead vehicle conditioned IMLE is shown in Algorithm~\ref{alg:imle_proc_fmstgccn}. The noise is added to the lead vehicle observed trajectory, it is a Gaussian white noise multiplied by a learning weight.
\begin{algorithm}[h]
\caption{Lead Vehicle Conditioned Implicit Maximum Likelihood Estimation (IMLE) Algorithm}
\label{alg:imle_proc_fmstgccn}
\begin{algorithmic}
\REQUIRE The dataset $D={(tr_{o}^i,\mathcal{V}_{\text{ego}}^i)}_{i=1}^K$ with $K$ samples and the model $\theta(.)$
\REQUIRE Define loss function $\mathcal{L}(.)$
\STATE Initialize the model
\FOR{$e = 1$ \textbf{to} $\text{Epochs}$}
    \STATE Pick a random batch $(tr_{o}, \mathcal{V}_{\text{ego}})$ from $D$
    \STATE Draw i.i.d. samples $\tilde{\mathcal{V}_{\text{ego}}}^{1},\ldots,\tilde{\mathcal{V}_{\text{ego}}}^{m}$ from $\theta(tr_{o})$ by adding the noise to the observation of the lead vehicle
    \STATE $\sigma(i) \gets \arg \min_{i} \mathcal{L}\left( \mathcal{V}_{\text{ego}}-\tilde{\mathcal{V}_{\text{ego}}}^{i}\right)\;\forall i \in m$
    \STATE $\theta \gets \theta - \eta \nabla_{\theta}\sigma(i)$
\ENDFOR
\STATE \textbf{Return:} $\theta$
\end{algorithmic}
\end{algorithm}
The loss function $\mathcal{L}$ is defined in Equation~\ref{eq:lossfunctionfmstgcnn}. The loss function comprises two parts. The first part is MSE between the closest predicted sample and the ground truth. The second part is a triplet loss where the anchor is the closest predicted sample, the positive example is the second closest generated sample, and the negative example is the furthest generated example from the ground truth.
\begin{multline}
    \mathcal{L}(\tilde{\mathcal{V}_{\text{ego}}}_{m\in M},\mathcal{V}_{\text{ego}}) = \lVert  \tilde{\mathcal{V}_{\text{ego}}}^\downarrow_1-\mathcal{V}_{\text{ego}} \rVert_2^2 \\+ \alpha (\lVert  \tilde{\mathcal{V}_{\text{ego}}}^\downarrow_1-\tilde{\mathcal{V}_{\text{ego}}}^\downarrow_2 \rVert_2^2 - \lVert  \tilde{\mathcal{V}_{\text{ego}}}^\downarrow_1-\tilde{\mathcal{V}_{\text{ego}}}^\downarrow_{-1} \rVert_2^2)
\label{eq:lossfunctionfmstgcnn}
\end{multline}
Where $M$ is the number of the generated samples, we set it to 20. The down arrow $\downarrow_r$ shows the proximity of the generated sample to ground truth, when $r=1$ it means the closest one and when $r = -1$ it means the furthest one. The $\alpha$ is weight for the triplet loss, we set it to $0.0001$. We train our model for 120 epochs starting from a learning rate of $0.001$ that drops each 40 epochs by a factor of $0.1$.
\section{Experiments \& Discussions}
In this section, we define the evaluation metrics used to evaluate the performance of our model. Then we define the baselines and compare our results against it. This follows with an ablation study of the \oursfm model design. We delve deeply by doing a qualitative analysis of our model and discuss how it reflects the data behavior. Last, we conclude this section with a qualitative analysis of the generated samples against the baseline models.
\subsection{Evaluation Metrics}
We use the commonly known metrics to evaluate the accuracy of the predicted results. Starting with the classic ones Average Displacement Error (ADE) and Final Displacement Error (FDE). The ADE measures the $L_2$ error between the predicted trajectory and the ground truth, defined in equation~\ref{eq:ADE}. The closer the prediction to the ground truth, the smaller the ADE is. The FDE measures if the prediction reached the last point of the ground truth defined in~\ref{eq:FDE}. Both ADE and FDE are deterministic metrics, when used to measure the accuracy of a generative model, the method of Best-of-N (BoN) is used. The BoN is to sample K samples, usually 20 samples, then choose the closest one to the ground truth and compute the ADE/FDE metrics based on this best sample. This makes the ADE/FDE insensitive to the whole distribution. For example, a model might have a low ADE/FDE because one of the tail samples is close to the ground truth. This is where the Average Mahalanobis Distance (AMD) and the Average Maximum Eigenvalues (AMV) metrics are used~\cite{mohamed2022social}. The AMD metric defined in equation~\ref{eq:AMD} measures the Mahalanobis distance between the ground truth point and the generated distribution. It measures the distance in terms of standard deviation units. Also, it enjoys analytical properties such as connection with $\chi^2$ distribution to have a reflection on the confidence intervals. The AMD can be very low, which is very good if the model has a huge uncertainty, thus the AMV metric is necessary to accompany the AMD metric. The AMV metric defined in equation~\ref{eq:AMV} measures the spread of the predicted distribution. The lower the AMD is the more certainty the model has. Thus, a suitable model will have low AMD/AMV metrics.
\vspace{-6pt}
\begin{equation}
\label{eq:ADE}
    \text{ADE} = \frac {1}{N \times T_{p}} \sum\limits_{n \in N} \sum\limits_{t \in T_{p}}\lVert \hat{p}^n_{\text{t}}-p^n_{\text{t}} \rVert_2
\end{equation}
\vspace{-6pt}
\begin{equation}
\label{eq:FDE}
    \text{FDE} = \frac {1}{N} \sum\limits_{n \in N}\lVert \hat{p}^n_{t}-p^n_{t} \rVert_2 , t = T_{p}
\end{equation}
\vspace{-6pt}
\begin{equation}
    \label{eq:AMD}
    \text{AMD} =
    \frac{1}{N \times T_p} \sum_{n \in N} \sum_{t \in T_p}  M_D\left(\hat{\mu}_{\text{GMM},t}^n,\hat{G}_t^n,p_t^n\right)
\end{equation}
\vspace{-6pt}
\begin{equation}
    \label{eq:AMV}
    \text{AMV} =
    \frac{1}{N \times T_p} \sum_{n \in N} \sum_{t \in T_p}
    \lambda^\downarrow_1(\hat{\Sigma}_{\text{GMM},t}^n)
\end{equation}
Where $p_t^n \in P$ is the ground truth position at time step $t \in T_p$ for agent $n \in N$, $\hat{p}$ is the predicted position, $M_D$ is the Mahalanobis distance, $\hat{\mu}_{\text{GMM},t}^n$ is the mean of the GMM of the predicted point, $\hat{G}_t^n$ is a weighted covariance matrix of the GMM components and $\lambda^\downarrow_1$ is the largest Eigenvalue of the GMM covariance matrix $\hat{\Sigma}_{\text{GMM},t}^n$.
\subsection{Baseline Models}
We chose two models that act as a baseline to compare the performance of our model. The first one is used in prior works~\cite{MCMB21} as a representation of classical model-based approaches. The second one is selected as a representation of interaction based approaches. Though more baselines can be added for comparison, it will be redundant as we already show the performance of two major motion prediction groups. Beside this, our goal is to propose a baseline model for the lead vehicle following a motion prediction problem.
\setlist{nosep,after=\vspace{\baselineskip}}
\begin{itemize}[noitemsep]
    \item Kalman Filter: We used a Kalman filter to predict the trajectory of the ego-vehicle. This baseline is like the one used in the work of~\cite{MCMB21}.
    \item Social-STGCNN~\cite{mohamed2020social}: It is a deep model that treats the motion prediction problem as a graph, end to end. It was one of the earliest methods that used graphs to model the motion prediction problem.
\end{itemize}
\vspace{-9pt}
\subsection{Performance against baselines}
We compare our model with the previously mentioned baselines. Table~\ref{tab:fmstgcnn_expresults} illustrates these empirical results. Overall, \oursfm outperforms the baselines on the three time horizons on the AMD/AMV metrics. This shows that most of the predicted trajectories already fall with a proximity to the ground truth. Though the ADE/FDE results of our model are better than the baselines in most of the cases, it is not the best indicator for the accuracy of the results, as discussed earlier. Looking into the Kalman filter results, we notice it has an over confident estimation because of the AMV metric with a very high AMD. This is because the Kalman filter baseline is a linear model and it suffers a hard time predicting the turns of the ego-vehicle. The Social-STGCNN model has a very low AMD but an extremely high AMV compared with other models. This indicates a high uncertainty, because the design of Social-STGCNN depends on the interaction between agents in the prediction goal. This is something that does not exist in the introduced problem, as modeling the interaction is needed on the input side while on the output side it is not. We also notice an expected behavior. As the prediction horizon increases from 3 seconds to 8 seconds into the future, the error in all the models also increases. In the qualitative section results, we visualise the generated samples to have more in-depth analysis of the model.

\begin{table*}[h]
\scriptsize
\centering
\caption[Performance of \oursfm versus baselines]{Performance of \oursfm versus baselines. The lower the better. Different predictions horizons are shown.}
\label{tab:fmstgcnn_expresults}
\begin{tabular}{|l|ll|ll|ll|}
\cline{1-7}
\multicolumn{1}{|l|}{Model/ Prediction Horizon} & \multicolumn{2}{c|}{3 Seconds}    & \multicolumn{2}{c|}{5 Seconds}  & \multicolumn{2}{c|}{8 Seconds}    \\
\multicolumn{1}{|l|}{} & AMD/AMV       & ADE/FDE    & AMD/AMV   & ADE/FDE    & AMD/AMV    & ADE/FDE      \\
\hline
Kalman Filter~\cite{MCMB21}         & 44.37/ 0.0086 & \textbf{4.23}/10.42 & 57.62/0.0234   & 9.23/23.47 & 74.71/0.0592   & \textbf{19.17}/48.74  \\
Social-STGCNN~\cite{mohamed2020social}         &   7.95/1639558               &  113.39/18.06          &   14.92/21480             &       61.82/14.38     &     15.25/67322          &       86.91/\textbf{16.35}       \\
FollowMe-STGCNN       &    \textbf{12.35/3.67}              &     4.57/\textbf{10.01}       & \textbf{30.01/49.54}   &    \textbf{7.67/16.05}        &      \textbf{53.44/11.98}         &   37.89/76.16           \\
\hline
\end{tabular}

\end{table*}

\begin{figure*}[ht]
    \centering
    \includegraphics[width=0.8\linewidth,height=3in,keepaspectratio]{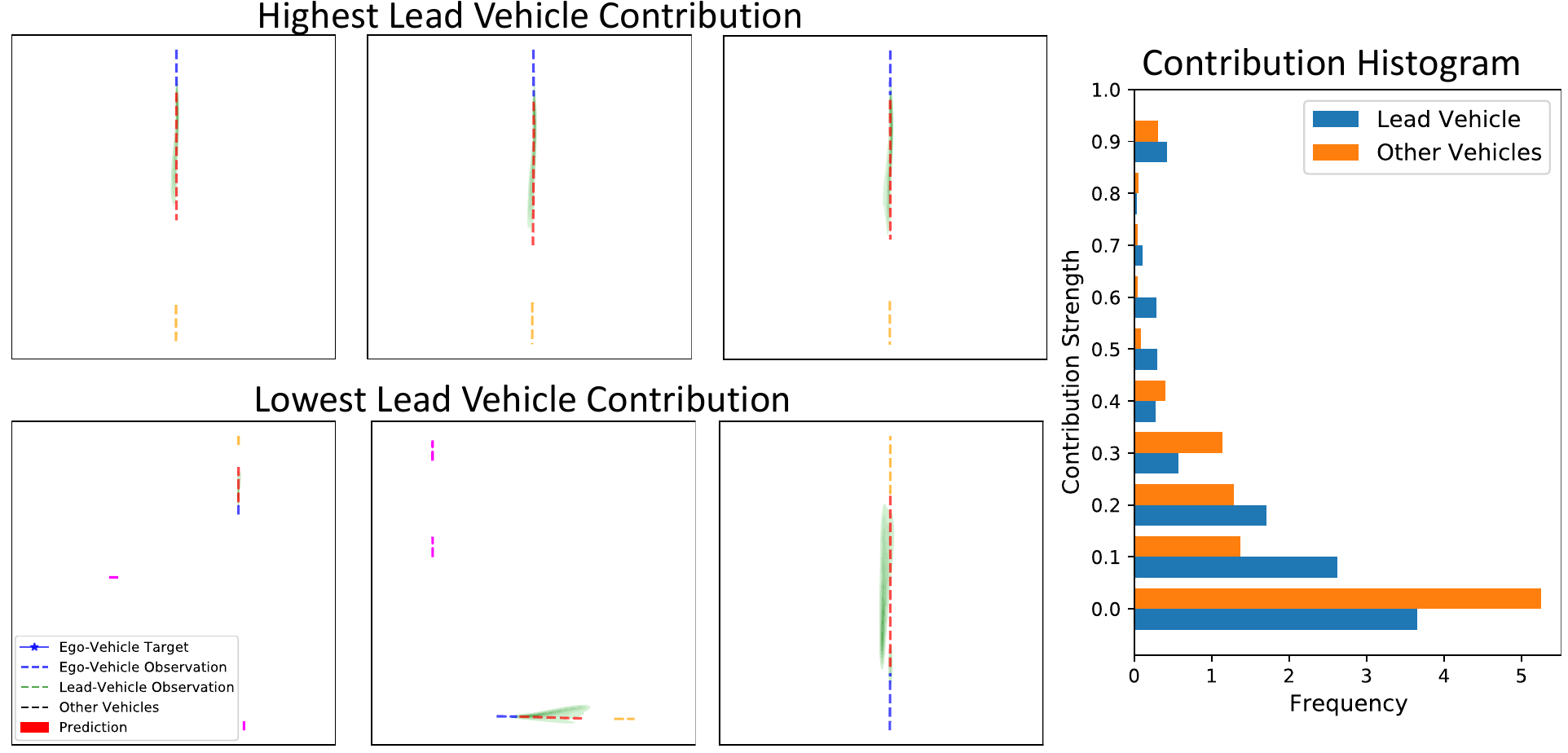}
    \caption[Contribution of lead vehicle versus the others vehicles]{Contribution of lead vehicle versus the others vehicles. The bar chart shows the histogram of contribution strength across the test dataset. The figures show different situations where the lead vehicle had the highest contributions and the lowest contributions.}
    \label{fig:contr_lead_vs_other}
\end{figure*}
\begin{figure}[ht]
    \centering
    \includegraphics[width=0.8\columnwidth,height=2in,keepaspectratio]{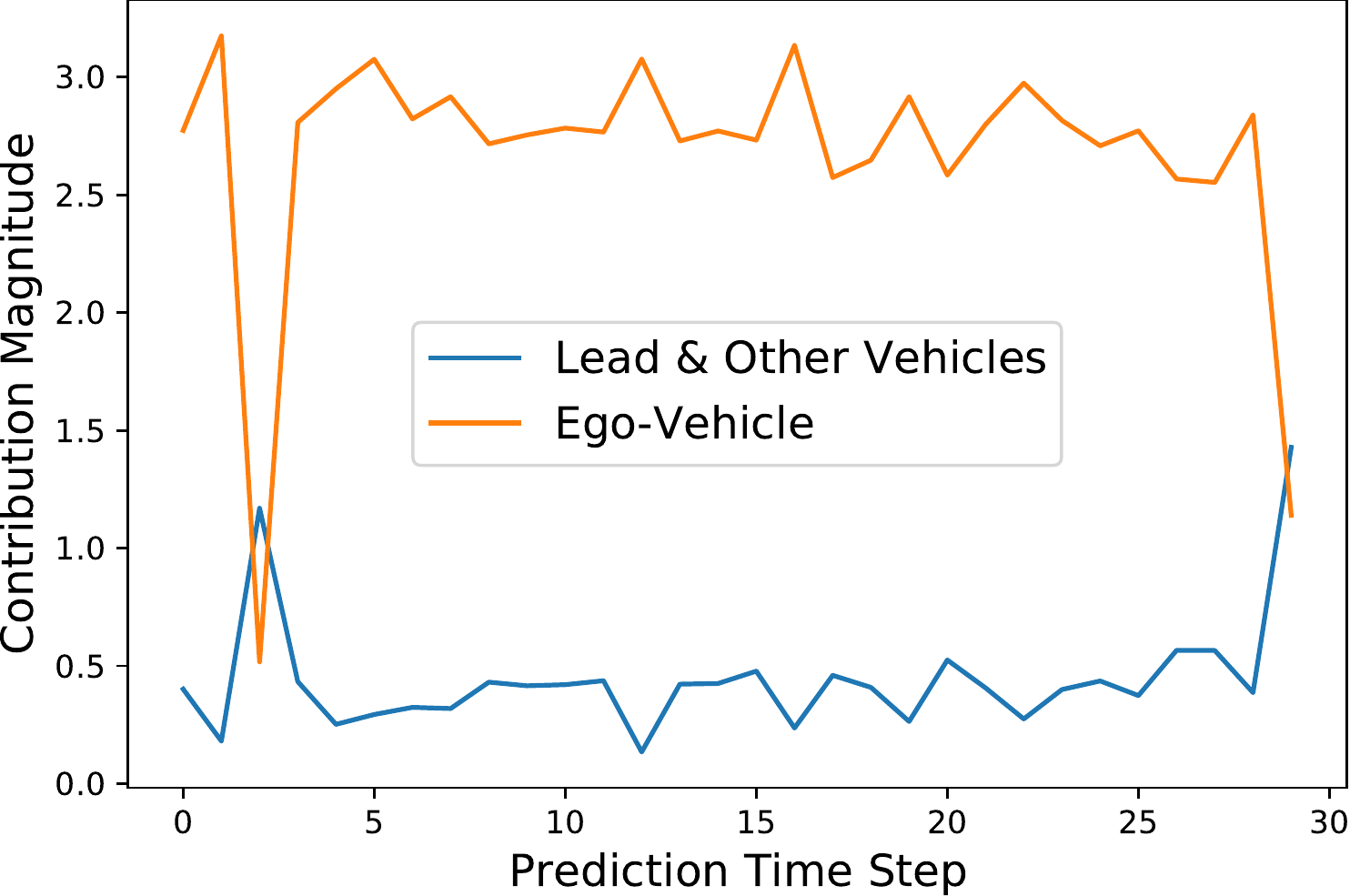}
    \caption[Contribution of lead and other vehicles versus the ego vehicle embedding]{Contribution of lead and other vehicles versus the ego vehicle embedding into predicting the trajectory of the ego vehicle.}
    \label{fig:contribstr}
\end{figure}

\subsection{\oursfm behavior analysis}
In order to better understand our model behavior, we developed two analyses. The first one measures the contribution of the lead vehicle history versus the other vehicles measured on the output of the fusion temporal 2D CNN, which is defined in Equation~\ref{eq:fusion_weight}. The contribution ranges between 0-1 and the value of this contribution is being multiplied by the embedding per trajectory. For example, a contribution of 0 means no embedding is being taken from this trajectory, while a contribution of 1 means the full embedding is being considered. In Figure~\ref{fig:contr_lead_vs_other} we show the contribution strength alongside the scenarios with high or low lead vehicle contribution. We notice from the bar chart that at high values of contribution bigger than 0.5 the lead vehicle contributes the most. This was expected because the driver is following the lead vehicle, so it has the most influence on the predicted trajectory. We also notice that the other vehicles might not influence the trajectory at all with the highest concentration at 0 contribution. When we look into the scenarios, the highest contribution of the lead vehicle indeed exists when there are no other vehicles around. But when other vehicles exist, the lead vehicle contribution is the lowest. This reflects the trust of the ego-vehicle driver in the lead vehicle, something that needs to be considered when the lead vehicle routes are being planned. Also, interestingly, in the bottom right case, the contribution of the lead vehicle is low because it is too close to the ego-vehicle. The existence of this behavior in our model fortifies the need for the fusion block and reflects real aspects of the data behavior that was captured by our model. The second behavior analysis measures the contribution magnitude from all lead and other vehicles embedding versus the contribution of the history of the ego vehicle. This is the analysis of the two operands of Equation~\ref{eq:final_output}. We can see this in Figure~\ref{fig:contribstr}. We notice on average the ego-vehicle trajectory has the highest influence on the final predicted trajectory except around start and the end of the predicted trajectory. This can be understood that the predicted trajectory needs a lot of information to determine how the predicted trajectory will be at the beginning of the prediction. The same information to determine the final predicted point is also needed.

\subsection{Qualitative analysis}
\begin{figure*}
\centering
\begin{tabular}{l}
\includegraphics[ width=\linewidth]{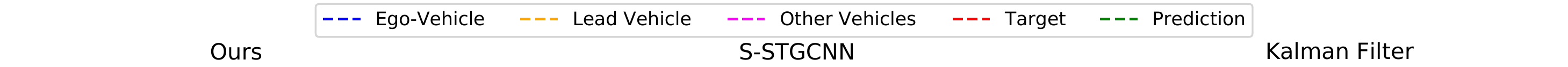}          \\
\includegraphics[ width=\linewidth]{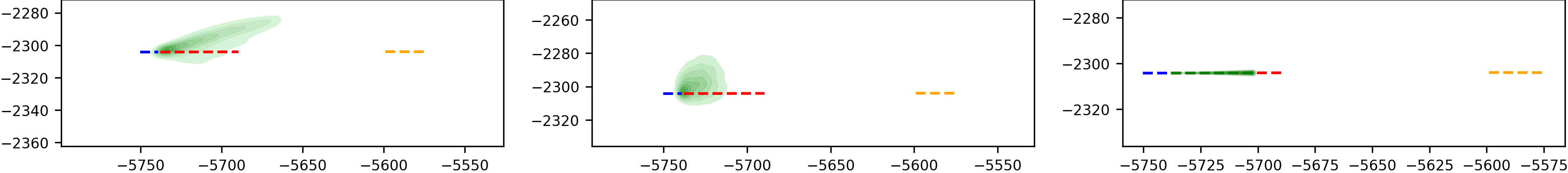}          \\
\includegraphics[ width=\linewidth, keepaspectratio]{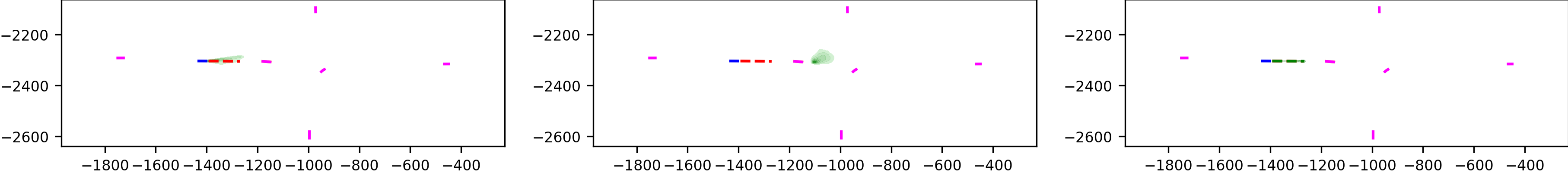}         \\
\includegraphics[ width=\linewidth, keepaspectratio]{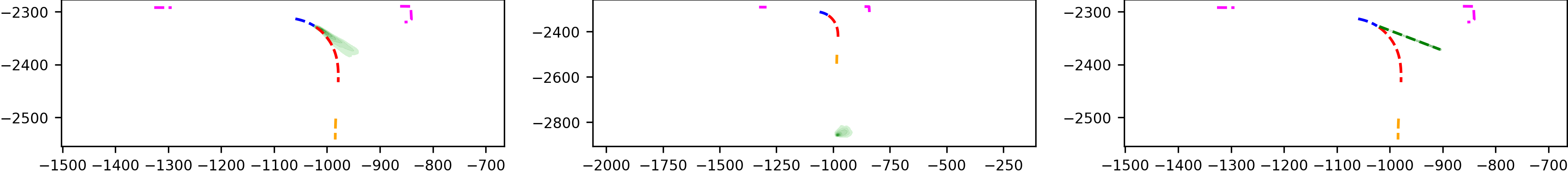}         \\
\midrule
\includegraphics[ width=\linewidth]{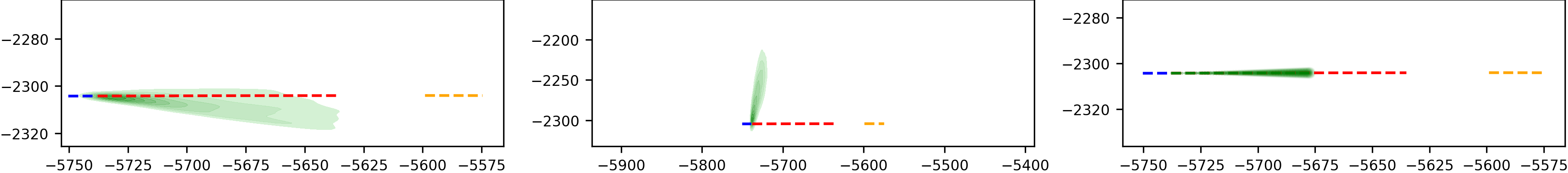}          \\
\includegraphics[ width=\linewidth, keepaspectratio]{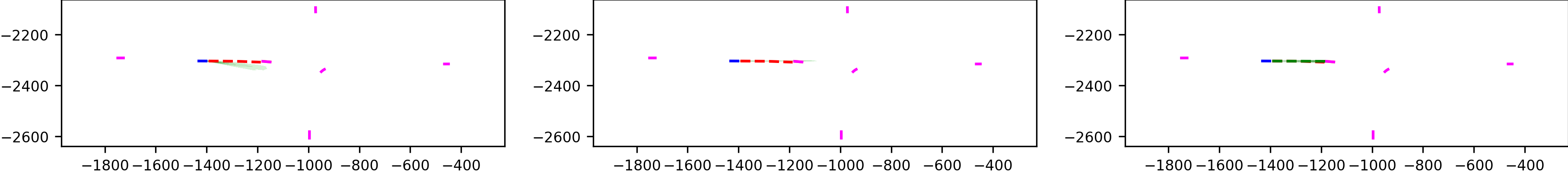}         \\
\includegraphics[ width=\linewidth, keepaspectratio]{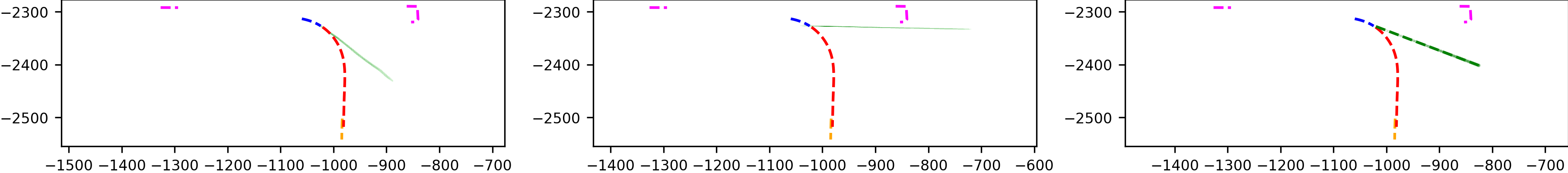}         \\
\midrule
\includegraphics[ width=\linewidth]{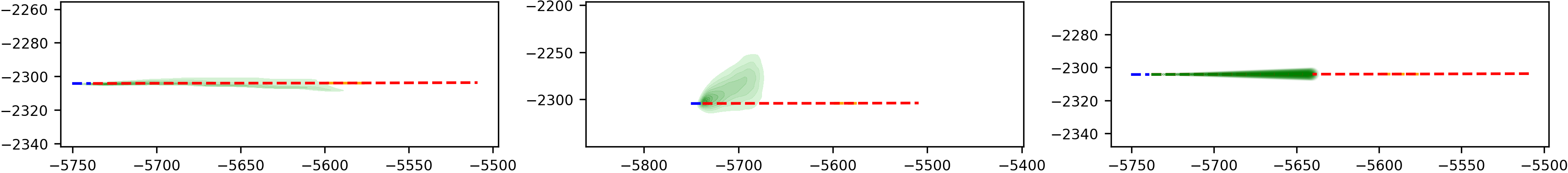}          \\
\includegraphics[ width=\linewidth, keepaspectratio]{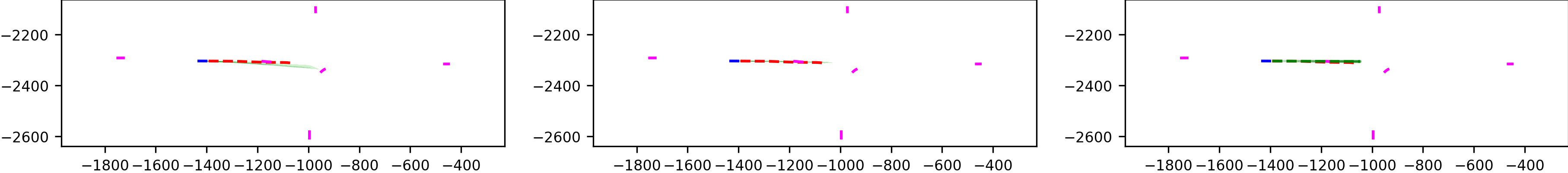}         \\
\includegraphics[ width=\linewidth, keepaspectratio]{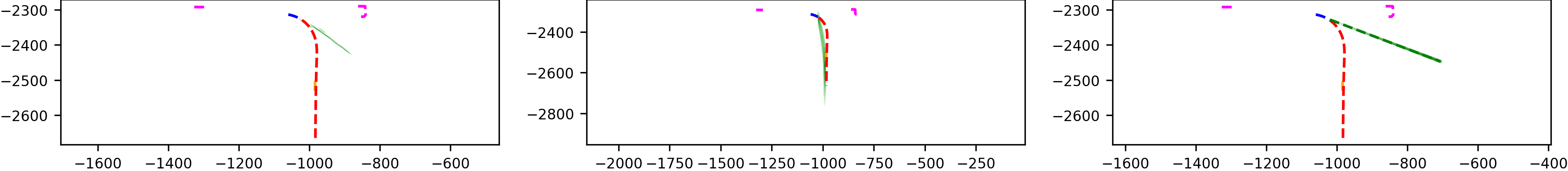}         \\
\end{tabular}
\captionof{figure}{Visualization of the predicted trajectories by several models. The three sections from top to bottom correspond to 3, 5 and 8 seconds of predictions into the future. Several cases are shown from easy to difficult scenarios.}
\label{fig:qual_analysis_fmstgcnn}
\end{figure*}

In this section, we visualize the prediction of our model and the baselines. We look into different situations such as situations with or without other vehicles, turning in intersections and different prediction horizons. Figure~\ref{fig:qual_analysis_fmstgcnn} illustrates different situations with three sections. The top one is the prediction of 3 seconds into the future. The middle and the last one are 5 and 8 seconds of prediction into the future. As a general observation, the Kalman filter is certain in the predictions, reflecting the results of the AMV metrics, but the predictions do not exhibit any non-linearity and are short when compared to the ground truth. The Social-STGCNN predictions are off the start of the prediction ground truth with over shooting or under shooting. This is because Social-STGCNN was tailored for social interaction between agents in the prediction step, something does not exist in our problem formulation. Last, our model \oursfm captures most of the situations to an extent with a clear bias towards the lead vehicle. The uncertainty of our model increases with the time step, but in most of the cases, it can extend the prediction to the full length of the ground truth, unlike the baselines. 

Delving into the predictions, over the short term of 3 seconds, our model clearly behaves better than the baselines in the different situations. We notice in the turning right case that our model is predicting the turn unlike the Kalman filter, which is straight and the Social-STGCNN which results in a weird prediction. In the 5 seconds case, our model extends the prediction into the expected ground truth while the others are short. Yet, our model fails to capture the full curve in the turning right case. While in the 8 seconds our model extends the predictions but not that far or shortens it similar to the Kalman filter situation. On the turning right case, both our model and the Kalman filter fails to capture it while the Social-STGCNN predicts the end correctly while ignoring the knob of the right turn. Overall, our model uses the information of the lead vehicle in a good manner, resulting in an overall good performance when compared to the baselines. This is because our design focuses only on predicting the ego vehicle trajectory and incorporating all other information from lead and other vehicles into the ego vehicle prediction.

\subsection{Ablation study}
There are two major design components in \oursfm model the fusion block and the triplet loss. The fusion block is the major component that defines the contribution between all other vehicles and the ego vehicle. It is used to enhance the prediction of the ego vehicle trajectory. The second component is the triplet loss, which forces the output space to create groups with similar samples and also encourages the model to reject odd samples. The triplet loss helps in minimizing the variance of the prediction, this can be reflected into the AMV metric. In Table~\ref{tab:fmstgcnn_ablation} we conduct an ablation study with these components. We notice the following, our model without the triplet and fusion block suffers from a huge AMD metrics. This means that the model is completely uncertain about the predictions as there is no guidance from the situation around the vehicles, also loose odd samples are not being regulated. When the fusion block is only used, we notice that the variance in the predicted trajectories is smaller than the previous case, reflecting the importance of this block. Yet, the AMD metrics went higher than the previous case, meaning that the generated samples went away from the ground truth. In order to achieve this balance between the AMD/AMV, when both the fusion and the triplet loss are being used we reach the lowest AMD/AMV metric. This means our predictions are close to the ground truth with more certainty. The ADE/FDE metrics only indicates the performance of the best sample. So, it does not reflect the whole generated distribution, unlike the AMD/AMV metrics. In the behavior analysis section, we ablate more the fusion block behavior.

\begin{table*}[ht]
\scriptsize
\centering
\caption[Ablation study of \oursfm components]{Ablation study of \oursfm components. The lower the better. The behavior of the fusion block and the triplet loss.}
\label{tab:fmstgcnn_ablation}
\begin{tabular}{|l|ll|ll|ll|}
\cline{1-7}
\multicolumn{1}{|l|}{Configuration/ Prediction Horizon}    & \multicolumn{2}{c|}{3 Seconds}     & \multicolumn{2}{c|}{5 Seconds}    & \multicolumn{2}{c|}{8 Seconds}       \\
\multicolumn{1}{|l|}{}    & AMD/AMV      & ADE/FDE    & AMD/AMV       & ADE/FDE    & AMD/AMV     & ADE/FDE       \\
\hline
Triplet + Fusion         & \textbf{12.35/3.67}    &4.57/10.01   &\textbf{30.01/49.54}     & \textbf{7.67/16.05}  & \textbf{53.44/11.98}     & 28.75/56.95   \\
w/o Triplet + Fusion     & 98.87/1.47    & 6.41/13.49 & 28.75/56.95   & 9.09/56.95 & 821.73/43.03 & 59.03/123.08  \\
w/o Triplet + w/o Fusion & 4.78/138.87                &   \textbf{4.50/8.60}        & 3.40/98.89                & \textbf{7.66/15.92}          & 37.23/55.56            & \textbf{17.86/41.80 }          \\
\hline
\end{tabular}
\end{table*}

\section{Conclusion}
We presented the FollowMe dataset that discusses the problem of following a lead vehicle in a heterogeneous environment of autonomous and non-autonomous vehicles. Our dataset was contrasted with previous datasets and showed it fills a gap in this area. We showed that drivers do not follow the given lead vehicle route, resulting in a motion prediction problem that is shaped to understand this pattern. We also introduced \oursfm, a deep spatio-temporal graph model that acts as a baseline for the dataset. This model is designed in a way that reflects the systematic bias in the dataset from following the lead vehicle. We compared our model with previous works and provided both quantitative and qualitative analyses to understand our model behavior.
\bibliography{ref}

\begin{thebibliography}{26}
\providecommand{\natexlab}[1]{#1}
\providecommand{\url}[1]{\texttt{#1}}
\expandafter\ifx\csname urlstyle\endcsname\relax
  \providecommand{\doi}[1]{doi: #1}\else
  \providecommand{\doi}{doi: \begingroup \urlstyle{rm}\Url}\fi

\bibitem[Akash et~al.(2020)Akash, Jain, and Misu]{akash2020toward}
Akash, K., Jain, N., and Misu, T.
\newblock Toward adaptive trust calibration for level 2 driving automation.
\newblock In \emph{Proceedings of the 2020 International Conference on
  Multimodal Interaction}, pp.\  538--547, 2020.

\bibitem[Alahi et~al.(2016)Alahi, Goel, Ramanathan, Robicquet, Fei-Fei, and
  Savarese]{alahi2016social}
Alahi, A., Goel, K., Ramanathan, V., Robicquet, A., Fei-Fei, L., and Savarese,
  S.
\newblock Social lstm: Human trajectory prediction in crowded spaces.
\newblock In \emph{Proceedings of the IEEE conference on computer vision and
  pattern recognition}, pp.\  961--971, 2016.

\bibitem[Awad et~al.(2018)Awad, Butt, Curtis, Lee, Fiscus, Godil, Joy, Delgado,
  Smeaton, Graham, et~al.]{awad2018trecvid}
Awad, G., Butt, A., Curtis, K., Lee, Y., Fiscus, J., Godil, A., Joy, D.,
  Delgado, A., Smeaton, A., Graham, Y., et~al.
\newblock Trecvid 2018: Benchmarking video activity detection, video captioning
  and matching, video storytelling linking and video search.
\newblock In \emph{Proceedings of TRECVID 2018}, 2018.

\bibitem[Ess et~al.(2008)Ess, Leibe, Schindler, , and van Gool]{eth_biwi_00534}
Ess, A., Leibe, B., Schindler, K., , and van Gool, L.
\newblock A mobile vision system for robust multi-person tracking.
\newblock In \emph{IEEE Conference on Computer Vision and Pattern Recognition
  (CVPR'08)}. IEEE Press, June 2008.

\bibitem[Ettinger et~al.(2021)Ettinger, Cheng, Caine, Liu, Zhao, Pradhan, Chai,
  Sapp, Qi, Zhou, et~al.]{ettinger2021large}
Ettinger, S., Cheng, S., Caine, B., Liu, C., Zhao, H., Pradhan, S., Chai, Y.,
  Sapp, B., Qi, C., Zhou, Y., et~al.
\newblock Large scale interactive motion forecasting for autonomous driving:
  The waymo open motion dataset.
\newblock \emph{arXiv preprint arXiv:2104.10133}, 2021.

\bibitem[Goodfellow et~al.(2014)Goodfellow, Pouget-Abadie, Mirza, Xu,
  Warde-Farley, Ozair, Courville, and Bengio]{goodfellow2014generative}
Goodfellow, I., Pouget-Abadie, J., Mirza, M., Xu, B., Warde-Farley, D., Ozair,
  S., Courville, A., and Bengio, Y.
\newblock Generative adversarial nets.
\newblock \emph{Advances in neural information processing systems}, 27, 2014.

\bibitem[Gupta et~al.(2018)Gupta, Johnson, Fei-Fei, Savarese, and
  Alahi]{gupta2018social}
Gupta, A., Johnson, J., Fei-Fei, L., Savarese, S., and Alahi, A.
\newblock Social gan: Socially acceptable trajectories with generative
  adversarial networks.
\newblock In \emph{Proceedings of the IEEE Conference on Computer Vision and
  Pattern Recognition}, pp.\  2255--2264, 2018.

\bibitem[Haddad \& Lam(2021)Haddad and Lam]{haddad2021self}
Haddad, S. and Lam, S.-K.
\newblock Self-growing spatial graph network for context-aware pedestrian
  trajectory prediction.
\newblock In \emph{2021 IEEE International Conference on Image Processing
  (ICIP)}, pp.\  1029--1033. IEEE, 2021.

\bibitem[Huang et~al.(2019)Huang, Bi, Li, Mao, and Wang]{huang2019stgat}
Huang, Y., Bi, H., Li, Z., Mao, T., and Wang, Z.
\newblock Stgat: Modeling spatial-temporal interactions for human trajectory
  prediction.
\newblock In \emph{Proceedings of the IEEE/CVF International Conference on
  Computer Vision}, pp.\  6272--6281, 2019.

\bibitem[Kingma \& Welling(2013)Kingma and Welling]{kingma2013auto}
Kingma, D.~P. and Welling, M.
\newblock Auto-encoding variational bayes.
\newblock \emph{arXiv preprint arXiv:1312.6114}, 2013.

\bibitem[Kipf \& Welling(2016)Kipf and Welling]{kipf2016semi}
Kipf, T.~N. and Welling, M.
\newblock Semi-supervised classification with graph convolutional networks.
\newblock \emph{arXiv preprint arXiv:1609.02907}, 2016.

\bibitem[Li et~al.(2020)Li, Yang, Tomizuka, and Choi]{li2020evolvegraph}
Li, J., Yang, F., Tomizuka, M., and Choi, C.
\newblock Evolvegraph: Multi-agent trajectory prediction with dynamic
  relational reasoning.
\newblock \emph{arXiv preprint arXiv:2003.13924}, 2020.

\bibitem[Li \& Malik(2018)Li and Malik]{li2018implicit}
Li, K. and Malik, J.
\newblock Implicit maximum likelihood estimation.
\newblock \emph{arXiv preprint arXiv:1809.09087}, 2018.

\bibitem[Liu et~al.(2021)Liu, Akash, Misu, and Wu]{liu2021clustering}
Liu, J., Akash, K., Misu, T., and Wu, X.
\newblock Clustering human trust dynamics for customized real-time prediction.
\newblock In \emph{2021 IEEE International Intelligent Transportation Systems
  Conference (ITSC)}, pp.\  1705--1712. IEEE, 2021.

\bibitem[Makansi et~al.(2021)Makansi, {\c{C}}i{\c{c}}ek, Marrakchi, and
  Brox]{MCMB21}
Makansi, O., {\c{C}}i{\c{c}}ek, {\"O}., Marrakchi, Y., and Brox, T.
\newblock On exposing the challenging long tail in future prediction of traffic
  actors.
\newblock In \emph{IEEE International Conference on Computer Vision (ICCV)},
  2021.
\newblock URL
  \url{http://lmb.informatik.uni-freiburg.de/Publications/2021/MCMB21}.

\bibitem[Mangalam et~al.(2020)Mangalam, Girase, Agarwal, Lee, Adeli, Malik, and
  Gaidon]{mangalam2020not}
Mangalam, K., Girase, H., Agarwal, S., Lee, K.-H., Adeli, E., Malik, J., and
  Gaidon, A.
\newblock It is not the journey but the destination: Endpoint conditioned
  trajectory prediction.
\newblock In \emph{European Conference on Computer Vision}, pp.\  759--776.
  Springer, 2020.

\bibitem[Mohamed et~al.(2020)Mohamed, Qian, Elhoseiny, and
  Claudel]{mohamed2020social}
Mohamed, A., Qian, K., Elhoseiny, M., and Claudel, C.
\newblock Social-stgcnn: A social spatio-temporal graph convolutional neural
  network for human trajectory prediction.
\newblock In \emph{Proceedings of the IEEE/CVF Conference on Computer Vision
  and Pattern Recognition}, pp.\  14424--14432, 2020.

\bibitem[Mohamed et~al.(2022)Mohamed, Zhu, Vu, Elhoseiny, and
  Claudel]{mohamed2022social}
Mohamed, A., Zhu, D., Vu, W., Elhoseiny, M., and Claudel, C.
\newblock Social-implicit: Rethinking trajectory prediction evaluation and the
  effectiveness of implicit maximum likelihood estimation.
\newblock In \emph{Computer Vision--ECCV 2022: 17th European Conference, Tel
  Aviv, Israel, October 23--27, 2022, Proceedings, Part XXII}, pp.\  463--479.
  Springer, 2022.

\bibitem[Salzmann et~al.(2020)Salzmann, Ivanovic, Chakravarty, and
  Pavone]{salzmann2020trajectron++}
Salzmann, T., Ivanovic, B., Chakravarty, P., and Pavone, M.
\newblock Trajectron++: Dynamically-feasible trajectory forecasting with
  heterogeneous data.
\newblock In \emph{Computer Vision--ECCV 2020: 16th European Conference,
  Glasgow, UK, August 23--28, 2020, Proceedings, Part XVIII 16}, pp.\
  683--700. Springer, 2020.

\bibitem[Sun et~al.(2020)Sun, Kretzschmar, Dotiwalla, Chouard, Patnaik, Tsui,
  Guo, Zhou, Chai, Caine, et~al.]{sun2020scalability}
Sun, P., Kretzschmar, H., Dotiwalla, X., Chouard, A., Patnaik, V., Tsui, P.,
  Guo, J., Zhou, Y., Chai, Y., Caine, B., et~al.
\newblock Scalability in perception for autonomous driving: Waymo open dataset.
\newblock In \emph{Proceedings of the IEEE/CVF Conference on Computer Vision
  and Pattern Recognition}, pp.\  2446--2454, 2020.

\bibitem[Tang et~al.(2020)Tang, Gui, Zhang, and Wang]{tang2020car}
Tang, T.-Q., Gui, Y., Zhang, J., and Wang, T.
\newblock Car-following model based on deep learning and markov theory.
\newblock \emph{Journal of Transportation Engineering, Part A: Systems},
  146\penalty0 (9):\penalty0 04020104, 2020.

\bibitem[{U.S. Department of Transportation Federal Highway Administration
  }(2016)]{NGSIM}
{U.S. Department of Transportation Federal Highway Administration }.
\newblock {Next Generation Simulation (NGSIM) Vehicle Trajectories and
  Supporting Data. Provided by ITS DataHub through Data.transportation.gov.
  Accessed 2022-01-12 from http://doi.org/10.21949/1504477}, 2016.

\bibitem[Wang et~al.(2020)Wang, Han, and Tiwari]{wang2020augmented}
Wang, Z., Han, K., and Tiwari, P.
\newblock Augmented reality-based advanced driver-assistance system for
  connected vehicles.
\newblock In \emph{2020 IEEE International Conference on Systems, Man, and
  Cybernetics (SMC)}, pp.\  752--759. IEEE, 2020.

\bibitem[Zhang et~al.(2019)Zhang, Sun, Qi, and Sun]{zhang2019simultaneous}
Zhang, X., Sun, J., Qi, X., and Sun, J.
\newblock Simultaneous modeling of car-following and lane-changing behaviors
  using deep learning.
\newblock \emph{Transportation research part C: emerging technologies},
  104:\penalty0 287--304, 2019.

\bibitem[Zhao \& Wildes(2021)Zhao and Wildes]{zhao2021you}
Zhao, H. and Wildes, R.~P.
\newblock Where are you heading? dynamic trajectory prediction with expert goal
  examples.
\newblock In \emph{Proceedings of the IEEE/CVF International Conference on
  Computer Vision}, pp.\  7629--7638, 2021.

\bibitem[Zhu et~al.(2018)Zhu, Wang, and Wang]{zhu2018human}
Zhu, M., Wang, X., and Wang, Y.
\newblock Human-like autonomous car-following model with deep reinforcement
  learning.
\newblock \emph{Transportation research part C: emerging technologies},
  97:\penalty0 348--368, 2018.

\end{thebibliography}
\bibliographystyle{icml2022}

\clearpage
\newpage
\appendix
\onecolumn

\end{document}